# FORWARD KINEMATICS SOLUTION FOR A GENERAL STEWART PLATFORM THROUGH ITERATION BASED SIMULATION

**Sourabh Karmakar**
PhD Student
Mechanical Engineering
Clemson University
Clemson, SC
skarmak@clemson.edu

**Cameron J. Turner**[1]
Associate Professor
Mechanical Engineering
Clemson University
Clemson, SC
cturne9@clemson.edu

**ACKNOWLEDGEMENTS**

The authors would like to acknowledge the support of Clemson University. All statements within are those of the authors and may or may not represent the views of these institutions.

---

[1] Corresponding Author

ABSTRACT

*This paper presents a method to generate feasible, unique forward-kinematic solutions for a general Stewart platform. This is done by using inverse kinematics to obtain valid workspace data and corresponding actuator lengths for the moving platform. For parallel kinematic machines, such as the Stewart Platform, inverse kinematics are straight forward, but the forward kinematics are complex and generates multiple solutions due to the closed loop structure of the kinematic links. In this research, a simple iterative algorithm has been used employing modified Denavit-Hartenberg convention. The outcome is encouraging as this method generates a single feasible forward kinematic solution for each valid pose with the solved DH parameters and unlike earlier forward kinematics solutions, this unique solution does not need to be manually verified. Therefore, the forward kinematic solutions can be used directly for further calculations without the need for manual pose verification. This capability is essential for the six degree of freedom materials testing system developed by the authors in their laboratory. The developed system is aimed at characterizing additively manufactured materials under complex combined multiple loading conditions. The material characterization is done by enabling high precision force control on the moving platform via in situ calibration of the as-built kinematics of the Stewart Gough Platform.*

# 1 INTRODUCTION

The Stewart Platform is one of the most popular parallel kinematic machines (PKMs) [1]. Though there are PKMs with 3, 4, 5, 6 parallel links, 6 parallel linked stationary PKMs are termed as Stewart Platform or Hexapod and are most versatile among the PKMs because of having six degrees of freedom available in a compact machine [2]. A typical hexapod consists of one fixed base plate and one movable plate or platform connected by six actuators. Each of the six parallel actuators generally provides one degree of freedom (DOF) to the machine by three translations along $x$, $y$, $z$ axes and three rotations about $x$, $y$, $z$ axes when a cartesian reference frame is attached to the movable platform center [3]. The hexapod considered here has each prismatic (P) actuator connected to the base with a universal (U) joint and the other end is connected to the moving platform by a spherical (U) joint. Such a hexapod or Stewart platform is designated as 6-UPS Parallel Kinematic Machine [4]. The movement of the platform in 6-DOF through the movement of six actuators makes the motion control complex compared to other machines.

To provide position and motion control, hexapod platforms use either Inverse Kinematics or Forward Kinematics. The position and orientation of the platform center point is known as *platform pose* [5]. In inverse kinematics the actuator lengths are calculated based on the platform pose [6]. In forward kinematics the platform pose is calculated for a given set of the actuator lengths and joint angles [7]. The mathematics and solution of inverse kinematics is much easier than forward kinematics for a hexapod. The complexity of forward kinematics, also called direct kinematics, is generated due to highly non-linear kinematic equations with multiple solutions. To solve the forward kinematics, many researchers tried different ways to solve the non-linear problem. A 16th-degree univariate polynomial on the 6-3 type PKM had been formulated by Innocenti and Parenti-Castelli [8]. Huang, Xiguang, Liao, Qizheng, at el. [9] presented algebraic method for a general 6-6 Stewart platform that yielded a 20th degree univariate polynomial from the determinant of the 15×15 Sylvester's matrix. Husty [10] derived a 40th-degree univariate equation for a general 6-6 Stewart platform, by finding the greatest common divisor of the intermediate polynomials. Domagoj & Leonardo Jelenkovicti [7] used canonical formulation for the forward kinematics and derived 9 equation with 9 unknowns and then solved it by multiple optimization methods. Wang, Yunfeng [11] derived the direct kinematics solutions for calculating the platform pose by increasing the actuator lengths in small amount and then increasing the joint parameters also in small amount utilizing numerical methods. Manuel Cardona [12] calculated the possible poses using Newton-Raphson Method for a set of joint angles and actuator lengths which includes invalid and valid poses for the platform and those need to be checked manually for acceptance. Also, other researchers [13], [14] used Newton-Rapson method to find the forward kinematic solution for parallel robots. The accuracy of convergence obtained in those works are at different levels based on the initial guesses. X. Zhou *et al.* [15] created pose error model with the help of Denavit-Hartenberg(DH) parameters and then converted it to a constrained quadratic optimization problem. In another research by M. Tarokh [16] used an approach to generate a lookup-table with possible solution space data. This solution space is divided into multiple clusters. For forward kinematic solution, the system looks at lookup-table clusters to get the required data directly or from the fitted curves with the available.

Other research on forward kinematics for Stewart platform used algebraic elimination [17], interval analysis, multiple optimization techniques, continuation algebraic formulations to generate solutions for a set of nonlinear equations or high degree of polynomials. Some researchers utilized neural network algorithms [18]–[21] and Artificial Intelligence (AI) [22] for improving the accuracy of the hexapod platform solving forward kinematics problem. All these works generated algorithms to obtain a valid solution for various types of PKMs, but finding a single feasible practical solution is still a challenging problem and limited for real-time applications.

The proposed method in this paper uses inverse kinematics to solve forward kinematics using modified Denavit-Hartenberg (DH) convention. DH convention is the most popular method for forward kinematics in serial manipulators, but its application remains very limited in parallel manipulators due to the closed loop nature of PKMs. The authors adopted this convention because of its simplicity and straight-forward nature of implementation. The proposed algorithm in this paper generates a single feasible orientation solution for a general Stewart Platform calculated by forward kinematics from a set of input data. The solution does not need to be validated through manual inspection. This is a simple iterative method using the available information from inverse kinematics. The pose data and corresponding actuator lengths are stored in a database. Then based on the motion limits of each joint, the algorithm generates the DH parameter dataset consisting of joint angles for the platform pose through iterative forward kinematics. The authors tried to exploit the power of the latest generation of computing systems by using simple iterative method which is not significantly more time-consuming method in the present days compared to the other efficient optimization methods like Newton-Rapson method, etc. Another advantage of this simple iterative method is that there is no initial guesses and no doubt about getting a solution (convergence). As long as a pose exists, a solution must be available.

The rest of the paper is organized into the following parts. The first part serves as an introduction. The next part elaborates the general mathematical expressions for a Stewart platform. Section 3 and 4 explain the inverse kinematics and forward kinematics used in the calculations. Section 5 explains the method that has been used in the algorithm to simulate the desired results for a Stewart platform-based Test frame "Tiger 66.1". Section 6 contains the result and discussion and finally the document ends with the conclusion.

## 2   PLATFORM POSE & WORKSPACE

Figure 1 shows the typical sketch of a hexapod platform. The circular plates at bottom and top are Fixed base and moving platform respectively. Two cartesian coordinate frames are attached at the center of each circular plate. The platform frame is defined by $P_x$, $P_y$ and $P_z$ axes with origin $O_P$ and the base coordinate frame is expressed by $B_x$, $B_y$ and $B_z$ axes with origin $O_B$. $P_z$ and $B_z$ are denoting the vertical axes of the respective frames. The orientation of the moving platform is defined by the orientation of the $O_P P_x P_y P_z$ frame with respect to base coordinate frame $O_B B_x B_y B_z$. The position of the platform center with respect to base center is defined by vector $h$. In one condition the base and platform frames remain parallel, actuators are at their smallest length and the z-axes are colinear. This orientation is called '*home pose*' [23]. In this condition, all the six actuators are of same length. Once the platform is moved by controlling the lengths of the actuators, the resultant motion at the center of the platform is a translation or rotation or a combination of both.

The position and orientation of the platform center depends on the values of roll, pitch, yaw, and the translation motion along $x$, $y$, $z$ axes as per the Euler angle representations [24]. The rotations are expressed by vector $\Phi$ and

$$\Phi = (\alpha\ \beta\ \gamma)^T \qquad (1)$$

where $\alpha$ (roll), $\beta$ (pitch), $\gamma$ (yaw) denotes the rotation angles about $x$, $y$ & $z$ axes respectively.

The translations are expressed by vector $d$ where

$$d = (dx\ dy\ dz)^T \qquad (2)$$

$dx$, $dy$ & $dz$ are the translation values along $x$, $y$ and $z$ axes respectively.

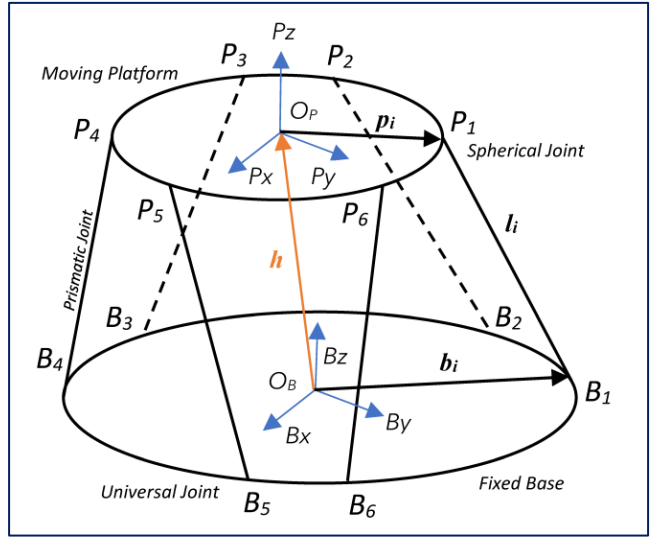

Figure 1: A typical hexapod configuration

The relationships between different platform poses and actuator variables are expressed by forward kinematics and inverse kinematics. In forward kinematics, the platform poses are calculated by the length and orientations of the six actuators. It can be expressed by Equation (3):

$$[dx, dy, dz, \alpha, \beta, \gamma]^T = f(q_1, q_2, q_3, \ldots\ldots, q_n) \tag{3}$$

where $dx, dy, dz, \alpha, \beta, \gamma$ are platform pose and $q_1, q_2, q_3 \ldots q_n$ are link variables that includes joint angles and actuator lengths.

In inverse kinematics, actuator lengths are calculated for a platform pose by Equation (4).

$$q_i = f_i(dx, dy, dz, \alpha, \beta, \gamma) \tag{4}$$

where $i = 1 \ldots n$ are link numbers. For a Stewart platform, the value for i = 1 to 6

Mathematically, the rotations about each axis are represented by the following equations as per Euler angle representation:

$$R_{x,\alpha} = \begin{bmatrix} 1 & 0 & 0 \\ 0 & c\alpha & -s\alpha \\ 0 & s\alpha & c\alpha \end{bmatrix} \tag{1a}$$

$$R_{y,\beta} = \begin{bmatrix} c\beta & 0 & s\beta \\ 0 & 1 & 0 \\ -s\beta & 0 & c\beta \end{bmatrix} \tag{5b}$$

$$R_{z,\gamma} = \begin{bmatrix} c\gamma & -s\gamma & 0 \\ s\gamma & c\gamma & 0 \\ 0 & 0 & 1 \end{bmatrix} \tag{5c}$$

Here *s* and *c* represent *sine* and *cosine* functions respectively.

Combining all these rotations, the complete rotation of the platform center with respect to its original fixed frame axes is calculated. The combined rotation denoted by **R** is calculated by pre-multiplying each subsequent rotation.

$$\boldsymbol{R} = R_{z,\gamma}.R_{y,\beta}.R_{x,\alpha}$$

$$= \begin{bmatrix} c\beta c\gamma & s\alpha s\beta c\gamma - c\alpha s\gamma & c\alpha s\beta c\gamma + s\alpha s\gamma \\ c\beta s\gamma & s\alpha s\beta s\gamma + c\alpha c\gamma & c\alpha s\beta s\gamma - s\alpha c\gamma \\ -s\beta & s\alpha c\beta & c\alpha c\beta \end{bmatrix} \tag{2}$$

Equation (2) represents the final rotation matrix. This is further combined with the *x*, *y*, *z* translations and the Homogeneous Transformation Matrix (HTM) [25] in Equation (3) is obtained. The complete HTM is expressed as

$$\boldsymbol{H} = \begin{pmatrix} \boldsymbol{R} & \boldsymbol{d} \\ 0 & 1 \end{pmatrix}$$

$$\therefore \boldsymbol{H} = \begin{bmatrix} c\beta c\gamma & s\alpha s\beta c\gamma - c\alpha s\gamma & c\alpha s\beta c\gamma + s\alpha s\gamma & dx \\ c\beta s\gamma & s\alpha s\beta s\gamma + c\alpha c\gamma & c\alpha s\beta s\gamma - s\alpha c\gamma & dy \\ -s\beta & s\alpha c\beta & c\alpha c\beta & dz \\ 0 & 0 & 0 & 1 \end{bmatrix} \tag{3}$$

## 3    INVERSE KINEMATICS & WORKSPACE

The platform pose with respect to the base center defines the state of the system. In inverse kinematics, the actuator lengths $l_i$ for a particular platform pose are calculated. With the change of vectors *Φ* and *d* the coordinate of the platform center $O_P$, platform joint vectors $p_i$, and vectors *h*, $l_i$ change. As the base is fixed, the base center vector $O_B$ and base joint vectors $b_i$ remain unchanged. The new values of $l_i$ and $h$ are calculated by calculating new values of $p_i$ and $O_P$ by finding value of $H$ in Equation (3) and pre-multiply to the respective coordinate values before the change. Then the length of each actuator is obtained by using Equation (4) [26]

$$l_i = ||h + {}^{B}R_P \cdot {}^{B}p_i - b_i||  \qquad (4)$$

where ${}^{B}R_P$ denotes the rotation vector to express the rotation of the platform coordinate frame with respect to the base coordinate frame, and ${}^{B}p_i$ is the platform joint vectors expressed with respect to the base coordinate frame.

For each PKM, the actuators operate between fixed length limits. Conforming to the values of $l_i$ for each new pose obtained by changing vectors $\Phi$ and $d$ define the feasible movable points for the platform center. All these feasible points form the *working space* for the selected hexapod platform.

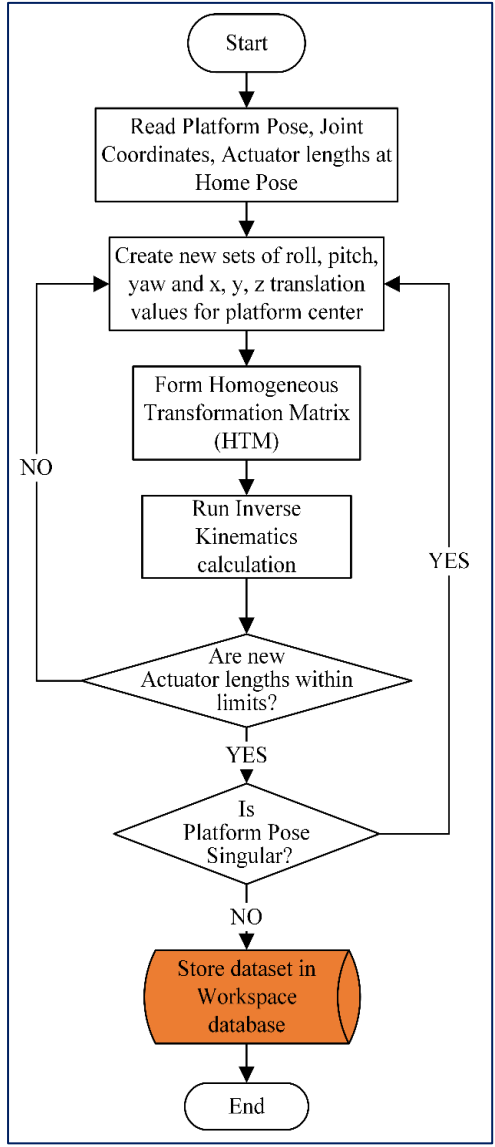
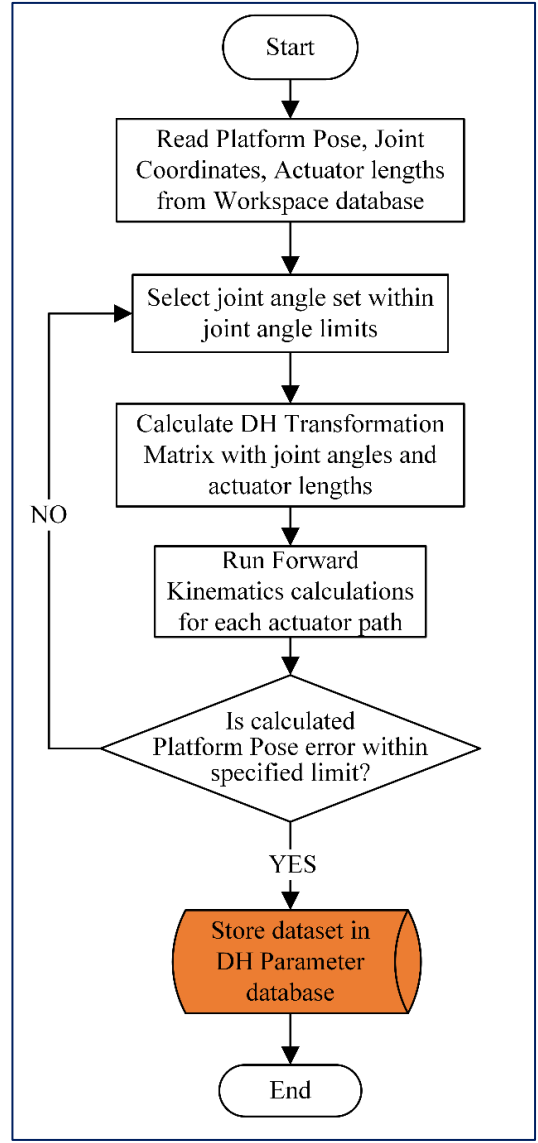

Figure 2: Flowchart for Workspace calculation by Inverse Kinematics

Figure 3: Flowchart for DH parameters calculation by Forward Kinematics

Figure 2 shows the flowchart of the inverse kinematics calculations used by the authors to generate the valid workspace points. An important aspect of the platform pose is *singularity*. In singular condition, the end-effector gains one or more unwanted instantaneous degrees of freedom (DOF) and the platform pose cannot be determined by unique actuator lengths. In such situation, the PKM becomes out of control and can transitorily make the drive force go infinity [27]. A pose with singularity is not considered as a valid workspace point and discarded from further calculations. The mathematical check for singularity is included inside the calculation code by checking if the determinant of the force Jacobian matrix in that pose is zero or not [28].

## 4 FORWARD KINEMATICS & VALID POSES

The Denavit-Hartenberg (DH) convention [29] is one of the most popular and earliest ways for solving forward kinematics for any serial manipulator in 3D space. For each pair of links, there are 4 DH parameters that are used in DH matrix to transfer the coordinate of a point from one coordinate frame to another.

The initial DH convention was introduced in 1955 by Jacques Denavit and Richard Hartenberg. In due course of time, it has been revisited by researchers and a modified DH convention [30] has been introduced. In this paper, modified DH convention has been used. The 4 DH parameters as per modified conventions are

Table 1: DH Parameters descriptions

| Modified DH Parameters | | | | |
|---|---|---|---|---|
| **Link ($i$)** | $a_{i-1}$ | $α_{i-1}$ | $r_i$ | $θ_i$ |
| Parameter name | Link length | Link twist | Link offset | Joint angle |

Here $i$ denotes the joint in consideration and $(i-1)$ is the previous joint. $i$ is always a positive integer. $^{i-1}T_i$ describes the transformation matrix for frame $i$ relative to frame $(i-1)$. Using these DH parameters, the frame transformation matrix is calculated by Equation (5) and (6) [31].

$$^{i-1}T_i = R_x(\alpha_{i-1}).T_x(a_{i-1}).R_z(\theta_i).T_z(r_i) \tag{5}$$

$$\textbf{Or,} \quad ^{i-1}T_i = \begin{bmatrix} cos\theta_i & -sin\theta_i & 0 & a_{i-1} \\ sin\theta_i cos\alpha_{i-1} & cos\theta_i cos\alpha_{i-1} & -sin\alpha_{i-1} & -r_i sin\alpha_{i-1} \\ sin\theta_i sin\alpha_{i-1} & cos\theta_i sin\alpha_{i-1} & cos\alpha_{i-1} & r_i cos\alpha_{i-1} \\ 0 & 0 & 0 & 1 \end{bmatrix} \tag{6}$$

If there are, suppose, 4 joints, they are numbered from 0 to 3. A point $P$ is expressed with respect to the end coordinate frame $\{3\}$ as $^3P = [P_x\ P_y\ P_z]^T$. The transformation matrix to express frame $\{3\}$ with respect to the base frame $\{0\}$ is calculated by Equation (7).

$$^0T_3 = \ ^0T_1 \cdot {^1T_2} \cdot {^2T_3} \tag{7}$$

And the point $^3P$ with respect to the base frame $\{0\}$ is calculated Equation (8).

$$^0P = \ ^0T_3 \cdot {^3P} \tag{8}$$

In the workspace calculation by inverse kinematics, the valid actuator lengths are calculated for each valid pose of the platform and these data are stored in the workspace database. For calculating the DH parameters, the MATLAB code follows the flowchart shown in Figure 3. The code iterates different combinations of joint angles $\theta_i$ to a reach to the pose coordinate using the corresponding actuator lengths available in the workspace database. The angles $\alpha_{i-1}$ are defined by the cartesian frame attached at each joints following modified DH convention rules and these values are not going to change for the whole process. A predefined error value is set in the calculation for the pose error which measures the distance between the pose in the database and pose calculated through forward kinematics. The forward kinematics calculation is repeated several times by changing $\theta_i$. Once the pose error is achieved below the predefined value, the DH parameter sets are stored in the DH parameter database for the pose. In this process there are some instances when the angle combinations cannot construct an orientation with a pose error below the defined error value, that pose is recorded without a valid DH parameter set.

## 5 SIMULATION RESULTS FOR "TIGER 66.1"

The above method of DH parameter calculation has been tried out through simulation for the Hexapod Test Frame "Tiger 66.1" developed by the authors in their lab. Tiger 66.1 is a special Stewart platform-based test frame developed for full-field characterization of additively manufactured specimens. A CAD model of the test frame is shown in Figure 4. A partial sketch also shows a couple of critical dimensions. The test process uses photogrammetry, so there are four cameras (2) mounted to capture data from the test zone. This restricts the movements of the platform center and overall workspace. The two green blocks on the upper part of the system are the grippers for holding the material specimen (4) to be tested. The upper gripper (1) is mounted on the fixed frame and the lower gripper (3) is fixed at the center of the hexapod top moving platform. All motions and forces are applied on the test specimen by moving the lower gripper.

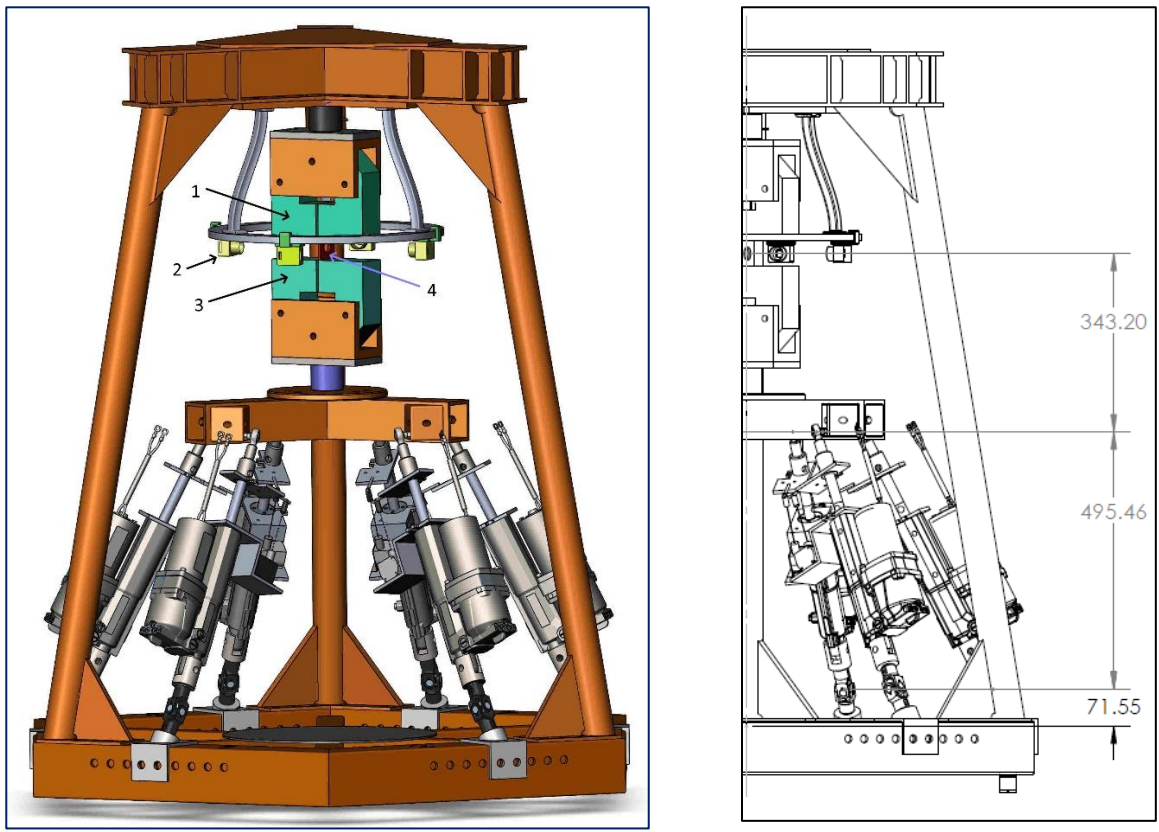

Figure 4: CAD model of hexapod Test Frame "Tiger 66.1" with partial sketch

Tiger 66.1 uses 6-UPS link combinations. As per the design and construction, the base frame center is situated 71.55 mm about the World frame origin which is located on the center of the circular plate at the lower structure and platform home pose is considered at a point 343.20 mm above the actual platform center point. This dimension does not change. The test frame lower and upper grip centers are considered at this point which is the condition of the system at the beginning of any test. All the calculations were done between the world frame center and this grip center pose. The major dimensions for Tiger 66.1 are shown in Table 2.

Table 2: Major dimensions of Tiger 66.1

| Sl no. | Parameters | Dimensions |
|---|---|---|
| 1 | Base joint circle radius | 477.4 mm |
| 2 | Smaller sides of base | 377.9 mm |
| 3 | Larger sides of base | 570.4 mm |
| 4 | Platform joint circle radius | 225.1 mm |
| 5 | Smaller sides of platform | 178.8 mm |
| 6 | Larger sides of platform | 268.7 mm |
| 7 | Actuator stroke length | 203.0 mm |
| 8 | Base center to Platform center height at test start | 495.46 mm |

The workspace for Tiger 66.1 has been calculated based on the actuator stroke length limit, translation, and rotation limits of the platform along and about $x$, $y$, $z$ axes respectively. The limits are given (Table 3) as per the feasible test conditions. All these motions limits are for pure translation and pure rotations.

Table 3: Motion limits of Tiger 66.1

| Motion no. | Parameters | Limits |
|---|---|---|
| 1 | Rotation about x-axis | +/- 30° |
| 2 | Rotation about y-axis | +/- 30° |
| 3 | Rotation about z-axis | +/- 55° |
| 4 | Translation along x-axis | +/- 158 mm |
| 5 | Translation along y-axis | +/- 158 mm |
| 6 | Translation along z-axis | 50 mm |

The workspace for the test frame is plotted and shown in Figure 5. Under given machine limits, the algorithm generated more than 180,000 valid workspace poses. This value may increase or decrease depending on the increment values used for platform's $dx, dy, dz, \alpha, \beta, \gamma$ parameters. The interval steps used in this case for translations along *x* & *y* axes = 15 mm, *z*-axis = 10 mm and +/-10° for rotations about each axis.

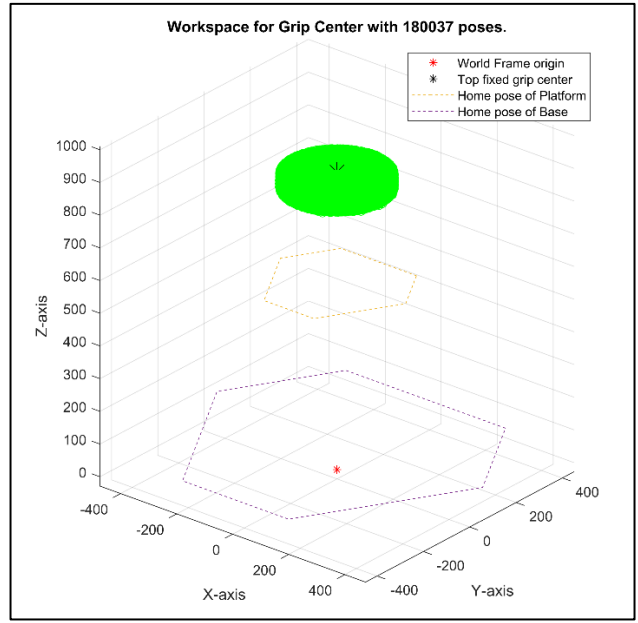

Figure 5: Graphical representation of Workspace

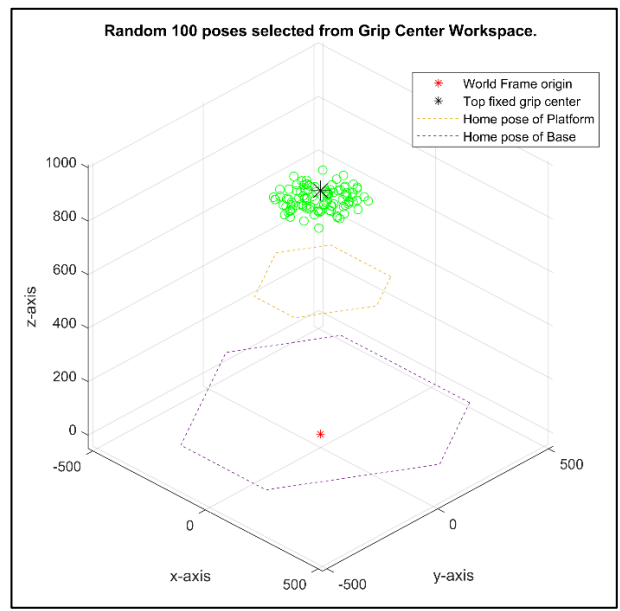

Figure 6: Selected 100 random poses from workspace

In the next part of the calculation, a random 100 poses have been selected (Figure 6) for finding the DH parameters for those poses through forward kinematics using DH transformation matrix.

For using the DH convention, coordinate frames have been assigned to each joint and a DH frame layout has been done (Figure 7).

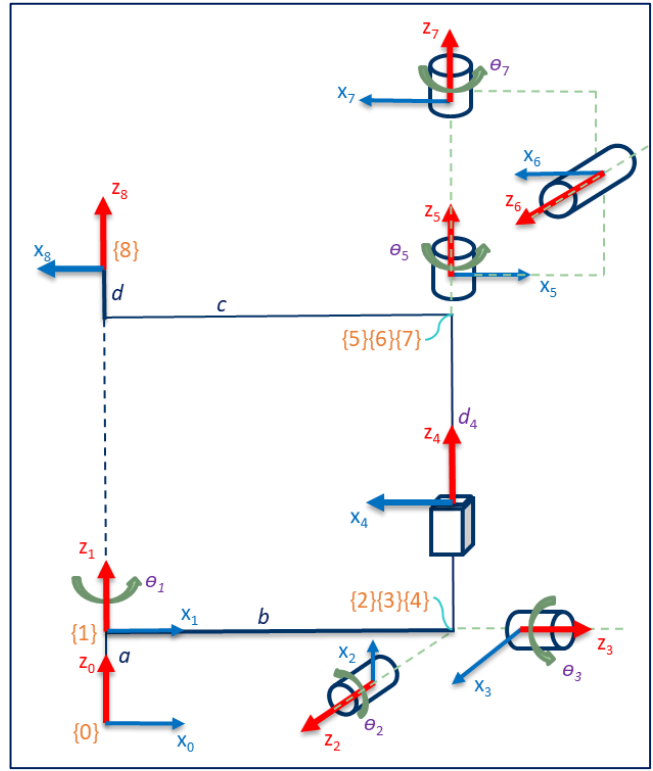

Figure 7: DH Frame layout for each actuator path

The layout shows the frames and variables assigned for one actuator path from world coordinate frame {0} to the moving grip center frame {8} on the platform. The coordinate frames were assigned as per the right-hand rule and modified DH convention. Frame {1} is the base center, frames {2} & {3} represent the universal joint at the bottom of each actuator, frame {4} denotes the prismatic joint on the actuator and frames {5}, {6}, {7} represent spherical joint at the top end of each actuator. As per construction, variables $a$, $b$, $c$, $d$ remain constant and represent height of the base center, base joint radius, platform joint radius and grip center height from the platform center respectively.

Table 4 lists the DH parameters for each actuator path. All joint angles are varying except $\theta_1$. The value of $\theta_1$ for each actuator path remains fixed due to the construction of Tiger 66.1.

Table 4: Modified DH parameter table

| Link (i) | $a_{i-1}$ | $\alpha_{i-1}$ | $r_i$ | $\theta_i$ |
|---|---|---|---|---|
| 1 | 0 | 0 | a | $\theta_1$ * |
| 2 | b | $\pi/2$ | 0 | $\theta_2$ |
| 3 | 0 | $\pi/2$ | 0 | $\theta_3$ |
| 4 | 0 | $\pi/2$ | 0 | $-\pi/2$ |
| 5 | 0 | 0 | d4 | $\theta_5$ |
| 6 | 0 | $\pi/2$ | 0 | $\theta_6$ |
| 7 | 0 | $\pi/2$ | 0 | $\theta_7$ |
| 8 | c | 0 | d | 0 |

* $\theta_1$ is predefined by construction for Tiger 66.1 & the values are 53.4°, 126.6°, 173.4°, 246.6°, 293.4°, 366.6°

The iterations found valid DH parameter sets for 100 poses out of 100 valid poses. To check the success rate of the algorithm, a random 100 poses were selected for 10 calculation processes and success rate of finding DH parameter sets is 100% for all processes. Table 5 shows the DH parameters value ranges as plotted in Figure 8 and the success of finding DH parameters for the number of poses considered.

Table 5: DH parameters value ranges and results obtained

| Variable | From | to |
|---|---|---|
| $\theta_2$ | -83.2° | -38.0° |
| $\theta_3$ | -118.8° | -63.4° |
| $\theta_5$ | 90.0° | 270.0° |
| $\theta_6$ | 92.0° | 231.0° |
| $\theta_7$ | -90.0° | 90.0° |
| $d_4$ | 465.68 mm | 664.68 mm |
| Number of poses evaluated | | 100 |
| Number of poses with valid DH parameters | | 100 |
| Number of poses with Grip Center deviation < 1mm | | 100 |

The DH parameter angles and actuator lengths variations for all 6 actuators for 100 poses are shown in Figure 8.

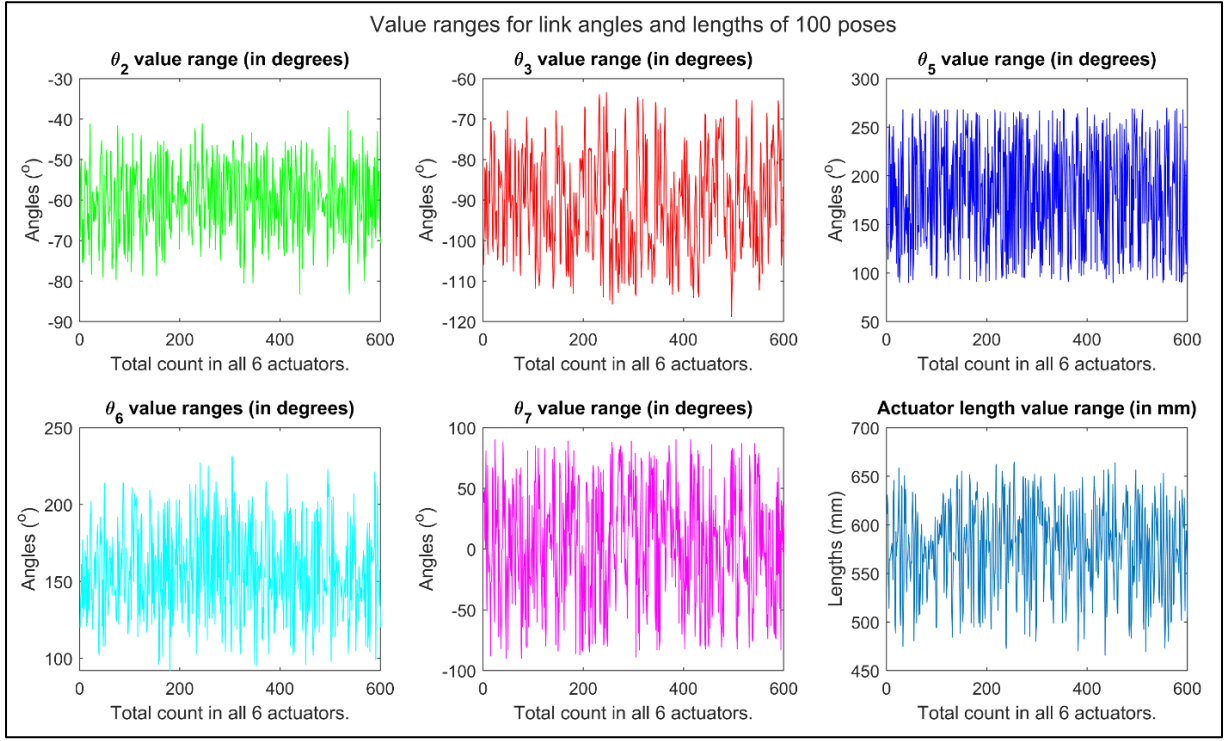

Figure 8: Angles and actuator length ranges

The orientations of Tiger 66.1 have been drawn based on the DH parameters obtained through the simulation. Figure shows random 6 numbers of such orientation layout and it is observed that all the orientations are valid and feasible. It has been verified that all DH parameters sets generate unique valid feasible poses every time, only some of them are shown here due to space limitations.

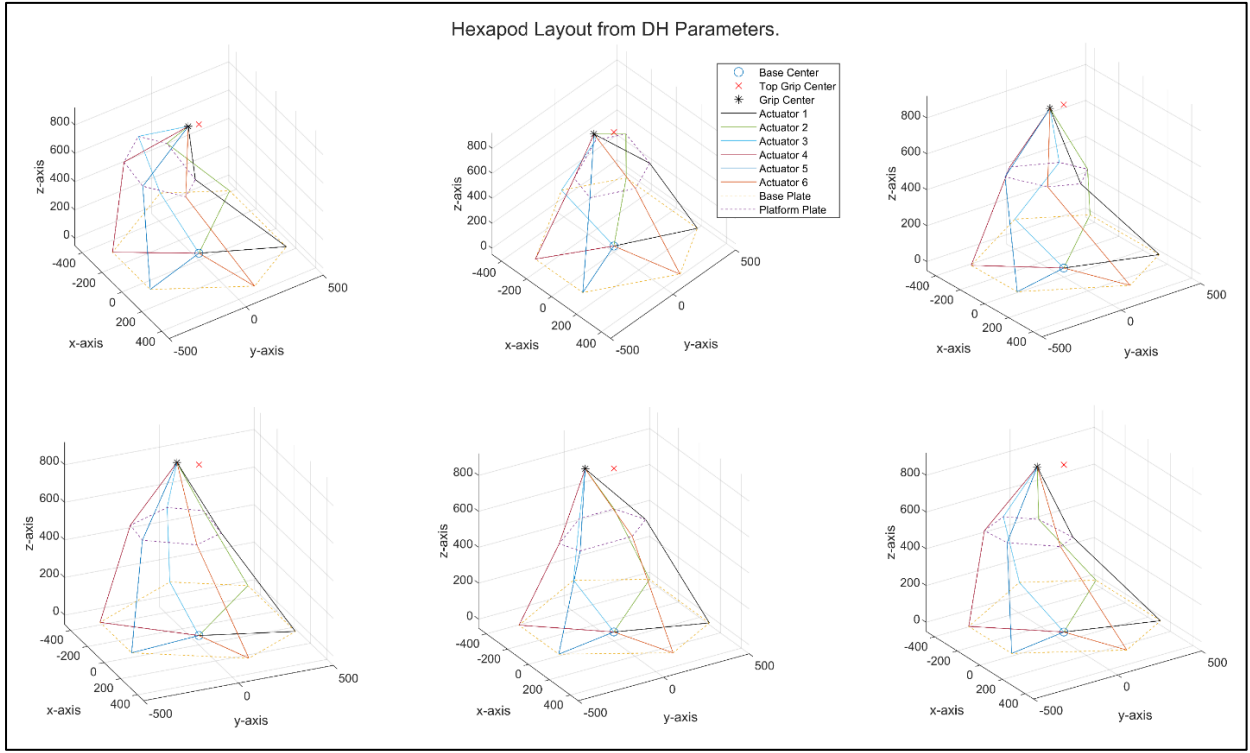

Figure 9: Tiger 66.1 orientation layouts through the calculated DH parameters

## 6 RESULTS AND DISCUSSION

The inverse kinematics and forward kinematics calculation run for Tiger 66.1 in this simulation shows that the unique solutions are achievable through the used algorithm. This algorithm does not use any complex calculations and found solutions through iterations. The pose data points were obtained for the workspace by changing the values of roll, pitch, yaw and translations along $x$, $y$, $z$ in each iteration. The number of data points is dependent on the increment step value of each variable. A smaller increment in value will generate more number of poses for the workspace. But the total time of calculation will increase because of an increased number of iterations.

The frame assignments from world frame to grip center frame at the joints through each actuator path can be done in different ways and the initial parameters values may be different.

This depends on whether the modified or standard DH conventions are used, but the result will be the same because both the DH transformation matrices generate the same unique solutions. For finding the DH parameters for a pose, the variable values are changed by predefined steps in each iteration. The finer the increment steps are, the more accuracy level is achieved, but the calculation time will increase. In the current simulation, DH parameters for 100% valid poses have been found with pose accuracy level < 1 mm. For these iterations the error limit for moving grip center pose was set as < 1 mm and the angle finding steps used between 0.1° to 1°. In a standard standalone laptop with Intel(R) Core(TM) i7-8550U CPU @ 1.80GHz and 16GB physical memory, it took on an average 32.09 seconds to solve the DH parameters for one pose. In any multicore modern server, this solution time can be reduced significantly. Trial showed that in a 56-core server, the execution time comes down to 1.28 seconds to get the same solution.

The orientations sketched in Figure from the solutions found through forward kinematics indicate that they are feasible and unique. The forward kinematics solutions done earlier by the researchers yielded more than one solution for one pose. Those solutions need to be inspected one by one and validated for the acceptable one. The forward kinematic results from the method explained in this paper do not require any manual review to check their feasibility. This gives an option to use the DH parameters for any further calculation without any intermediate stop and manual intervention for the selection of correct results. In section 6.1, we execute one such exercise which is important for the use of Tiger 66.1 in characterization of additively manufactured materials.

## 6.1 Pose deviation due to tolerances & Sensitivity

A real-world Stewart platform is not free from manufacturing and assembly tolerances. These errors cause the actual platform-pose to deviate from the theoretical platform pose for a set of DH parameters. The actual measurement of those DH parameter deviations is not only difficult and time consuming, but also expensive due to proper instrumentation. With various combinations of DH parameter tolerances, a tentative pose deviation can be calculated with this algorithm.

Five tolerance values were considered for calculating the pose error. These values are ±0.1, ±0.2, ±0.3, ±0.4 and ±0.5 in degrees for angles and in mm for the actuator lengths. These tolerances were applied to the DH parameters for 100 poses found through the calculations as discussed in the previous section. For ±0.5 deg & mm tolerances on the DH parameters, the platform-pose errors in terms of absolute *x*, *y*, *z* values and absolute distance values from the theoretical poses are shown in Figure 8. The 0 marked horizontal lines in both the plots indicate the theoretical values.

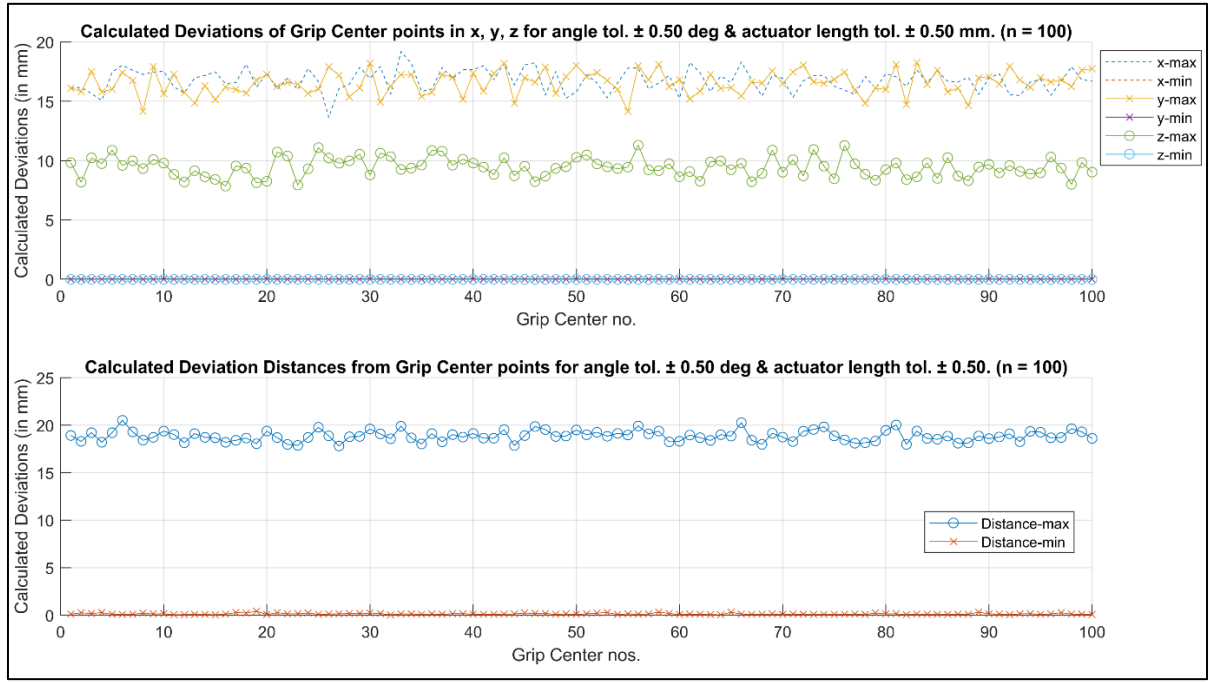

Figure 8: Platform pose deviations for DH parameters tolerances of ±0.5 deg & mm

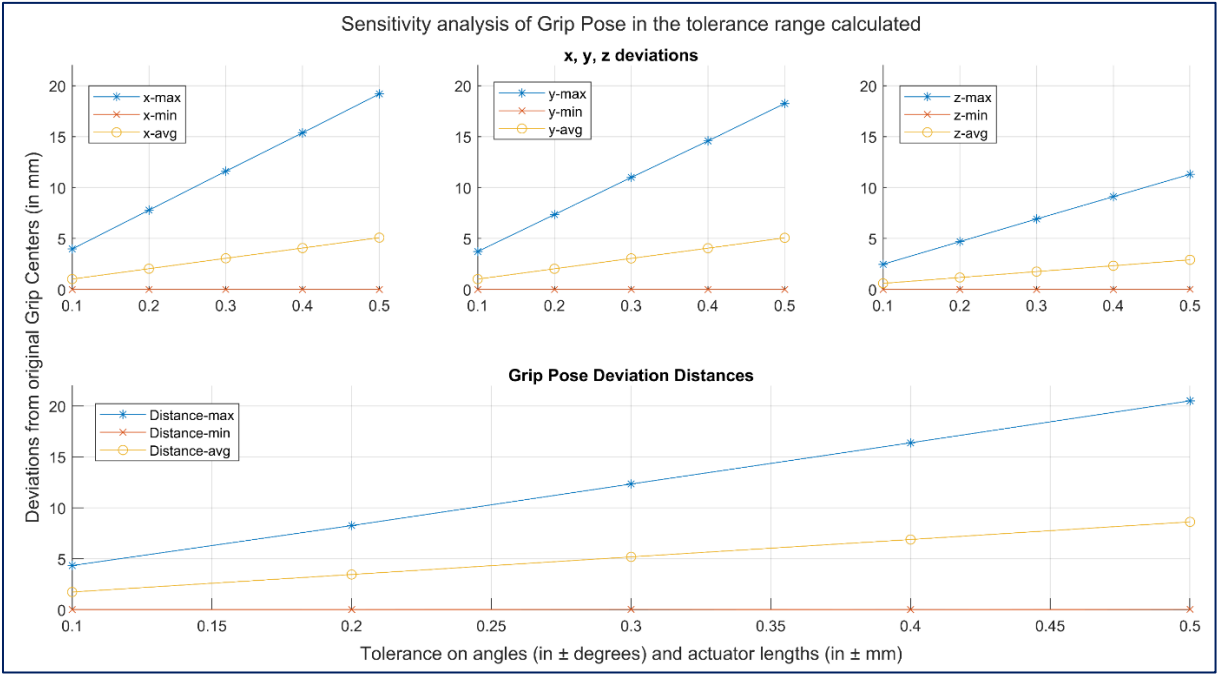

Figure 9: Sensitivity of Platform poses due to DH parameters tolerances

Similar calculations were done with four other tolerance values. From calculated data the maximum and minimum deviations were plotted to check the sensitivity of the platform poses due to the DH parameters tolerances (Figure 9). The maximum and average deviations for each tolerance were shown here. In all these cases, the Grip pose denotes the lower moving grip center mounted on the moving platform.

As seen from the plots, the maximum grip center or grip pose deviation occurs when tolerance value is ±0.5 for the DH parameters. The pose deviation distance may go more than 20 mm. This is of course a worst-case scenario where all the tolerance errors constructively combine to magnify the positional error in the end-effector position/orientation. It is far more likely that some tolerance increases the error value, while others decrease the error values. The deviations are appearing as linearly changing with the change of the tolerances, through that has not be verified in this work; but one more point has been found that the maximum and minimum values for the *x*, *y*, *z* coordinates as well as for the grip center deviations are not coincident. A summary of the observed data has been shown in Table 6.

Table 6: Pose deviations corresponding to *x*, *y*, *z* maximum & minimum values

| Tol. value | At | x-deviation | y-deviation | z-deviation | Grip Center deviation | At | x-deviation | y-deviation | z-deviation | Grip Center deviation |
|---|---|---|---|---|---|---|---|---|---|---|
| ±0.1 | x-max | **3.96** | 0.32 | 1.01 | **4.10** | x-min | **0.00** | 2.23 | 1.10 | 2.49 |
|  | y-max | 0.35 | **3.72** | 0.42 | 3.76 | y-min | 0.43 | **0.00** | 0.14 | **0.45** |
|  | z-max | 1.75 | 1.99 | **2.47** | 3.62 | z-min | 0.63 | 1.96 | **0.00** | 2.06 |
| ±0.2 | x-max | **7.78** | 0.79 | 1.95 | **8.06** | x-min | **0.00** | 1.18 | 0.60 | 1.33 |
|  | y-max | 1.59 | **7.35** | 1.40 | 7.65 | y-min | 1.82 | **0.00** | 0.52 | 1.89 |
|  | z-max | 4.80 | 3.32 | **4.69** | 7.49 | z-min | 0.96 | 0.38 | **0.00** | 1.03 |
| ±0.3 | x-max | **11.60** | 1.28 | 2.88 | **12.02** | x-min | **0.00** | 6.59 | 1.68 | 6.81 |
|  | y-max | 2.39 | **10.99** | 2.05 | 11.43 | y-min | 1.82 | **0.00** | 0.52 | 1.89 |
|  | z-max | 4.80 | 3.32 | **4.69** | 7.49 | z-min | 0.96 | 0.38 | **0.00** | 1.03 |
| ±0.4 | x-max | **15.37** | 1.79 | 3.79 | **15.93** | x-min | **0.00** | 1.69 | 0.27 | **1.71** |
|  | y-max | 2.55 | **14.57** | 4.60 | 15.49 | y-min | 4.18 | **0.00** | 1.79 | 4.54 |
|  | z-max | 9.56 | 6.65 | **9.10** | 14.78 | z-min | 4.10 | 4.47 | **0.00** | 6.07 |
| ±0.5 | x-max | **19.19** | 2.33 | 4.69 | **19.89** | x-min | **0.00** | 7.78 | 4.03 | 8.76 |
|  | y-max | 3.11 | **18.25** | 5.71 | 19.38 | y-min | 4.55 | **0.00** | 2.01 | 4.97 |
|  | z-max | 11.95 | 8.34 | **11.30** | 18.44 | z-min | 0.60 | 0.15 | **0.00** | **0.62** |

The bold lettered values in the above table represent the maximum and minimum values.

## 7 CONCLUSION

The simulation has been successfully executed for the test-frame Tiger 66.1. Both inverse and forward kinematics are completed with the iteration-based algorithms discussed here. The results indicate that the implementation of the algorithm for real time calculations is sensible. Once the valid workspace data and corresponding DH parameters for the poses are calculated, they can be stored in a database and those data can be used for real time applications. If a pose-data and related DH parameters are not available in the database, it can be calculated and added to the database to enrich it during the operation and can cover the whole workspace with more precise data. Also, the algorithm may be refined by reducing the iteration step values to generate valid DH parameter data sets for 100% valid poses with stricter error limits.

In this work, the authors ran the simulations with randomly selected 100 pose points. As the size of the database can grow bigger, finding the DH parameter data runtime for a required pose becomes trivial. The algorithm can be refined by adding efficient data searching methods.

There is further scope to validate these data with measurements done on the physical system for practical purposes. The authors intend to do this validation in their Tiger 66.1 hexapod platform in the next phase of this research through non-invasive methods like Photogrammetry. In photogrammetry the final platform pose can be measured directly without instrumenting and measuring each joint. By doing multiple actual measurements and comparing the values with the calculated values, the in-built construction or fabrication errors of system can be evaluated and applying proper compensation factors for those in-built deviations, the hexapod platform can be guided to the desired pose more accurately. Such a calibration process of a Stewart platform would make practical application of the machine more useful.

**FUNDING**

The authors express their appreciation to Clemson University who financially supported this work and to the United States Naval Research Laboratory in Washington, D.C. who provided technical insight and support for the design and operation of the TIGER 66.1 system through NCRADA-NRL-20-719.

**COMPETING INTERESTS**

The authors have no relevant financial or non-financial interests to disclose.

## AUTHOR CONTRIBUTIONS

All authors contributed to the conception and design of the research. Software coding, data collection and analysis, and preparation of the first draft were performed by Sourabh Karmakar. All authors commented on and contributed to previous manuscript versions. All authors read and approved the final manuscript.

## REFERENCES

[1]  B. Dasgupta and T. S. Mruthyunjaya, "A constructive predictor-corrector algorithm for the direct position kinematics problem for a general 6-6 Stewart platform," *Mech. Mach. Theory*, vol. 31, no. 6, pp. 799–811, 1996, doi: 10.1016/0094-114X(95)00106-9.

[2]  W. Tanaka, T. Arai, K. Inoue, Y. Mae, and C. S. Park, "Simplified kinematic calibration for a class of parallel mechanism," *Proc. - IEEE Int. Conf. Robot. Autom.*, vol. 1, no. May, pp. 483–488, 2002, doi: 10.1109/ROBOT.2002.1013406.

[3]  D. Daney, I. Z. Emiris, Y. Papegay, E. Tsigaridas, and J. P. Merlet, "Calibration of parallel robots : on the Elimination of Pose-Dependent Parameters," *EuCoMeS 2006 - 1st Eur. Conf. Mech. Sci. Conf. Proc.*, pp. 1–12, 2006.

[4]  A. C. Majarena, J. Santolaria, D. Samper, and J. J. Aguilar, "An overview of kinematic and calibration models using internal/external sensors or constraints to improve the behavior of spatial parallel mechanisms," *Sensors (Switzerland)*, vol. 10, no. 11, pp. 10256–10297, 2010, doi: 10.3390/s101110256.

[5]  B. Dasgupta and T. S. Mruthyunjaya, "Closed-form dynamic equations of the general Stewart platform through the Newton-Euler approach," *Mech. Mach. Theory*, vol. 33, no. 7, pp. 993–1012, 1998, doi: 10.1016/S0094-114X(97)00087-6.

[6]  J.-P. Merlet, "Still a long way to go on the road for parallel mechanisms," *Asme*, vol. 64,

no. May, pp. 1–19, 2002.

[7] D. Jakobović and L. Jelenković, "The forward and inverse kinematics problems for stewart parallel mechanisms," *CIM 2002 Comput. Integr. Manuf. High Speed Mach. - 8th Int. Sci. Conf. Prod. Eng.*, 2002.

[8] C. Innocenti and V. Parenti-Castelli, "Direct position analysis of the Stewart platform mechanism," *Mech. Mach. Theory*, vol. 25, no. 6, pp. 611–621, 1990, doi: 10.1016/0094-114X(90)90004-4.

[9] X. Huang, Q. Liao, S. Wei, X. Qiang, and S. Huang, "Forward kinematics of the 6-6 Stewart platform with planar base and platform using algebraic elimination," *Proc. IEEE Int. Conf. Autom. Logist. ICAL 2007*, no. 104043, pp. 2655–2659, 2007, doi: 10.1109/ICAL.2007.4339029.

[10] M. L. Husty, "An algorithm for solving the direct kinematics of general Stewart-Gough platforms," *Mech. Mach. Theory*, vol. 31, no. 4, pp. 365–379, 1996, doi: 10.1016/0094-114X(95)00091-C.

[11] Y. Wang, "A direct numerical solution to forward kinematics of general Stewart-Gough platforms," *Robotica*, vol. 25, no. 1, pp. 121–128, 2007, doi: 10.1017/S0263574706003080.

[12] M. Cardona, "Kinematics and Jacobian analysis of a 6UPS Stewart-Gough platform," *2016 IEEE 36th Cent. Am. Panama Conv. CONCAPAN 2016*, 2016, doi: 10.1109/CONCAPAN.2016.7942377.

[13] H. Zhang, Q. Gao, M. Zhang, and Y. Yao, "Forward kinematics of parallel robot based on neural network Newton-Raphson iterative algorithm," no. December 2021, p. 36, 2021, doi: 10.1117/12.2625254.

[14] C. Yang, Q. Huang, P. O. Ogbobe, and J. Han, "Forward kinematics analysis of parallel robots using global Newton-Raphson method," *2009 2nd Int. Conf. Intell. Comput. Technol. Autom. ICICTA 2009*, vol. 3, pp. 407–410, 2009, doi: 10.1109/ICICTA.2009.564.

[15] X. Zhou, F. Zhou, and Y. Wang, "Pose Error Modeling and Analysis for 6-DOF Stewart Platform," 2017.

[16] M. Tarokh, "Real time forward kinematics solutions for general Stewart platforms," *Proc. - IEEE Int. Conf. Robot. Autom.*, no. April, pp. 901–906, 2007, doi: 10.1109/ROBOT.2007.363100.

[17] J. M. Porta and F. Thomas, "Yet Another Approach to the Gough-Stewart Platform Forward Kinematics," *Proc. - IEEE Int. Conf. Robot. Autom.*, pp. 974–980, 2018, doi: 10.1109/ICRA.2018.8460900.

[18] R. KumarP and and BBandyopadhyay, *The Forward Kinematic Modeling of a Stewart Platform using NLARX Model with Wavelet Network*. 2013. doi: 10.1109/INDIN.2013.6622907.

[19] N. Fazenda, E. Lubrano, S. Rossopoulos, and R. Clavel, "Calibration of the 6 DOF high-precision flexure parallel robot 'Sigma 6,'" *Proc. 5th Parallel Kinemat. Semin. Chemnitz, Ger.*, pp. 379–398, 2006, [Online]. Available: http://scholar.google.com/scholar?hl=en&btnG=Search&q=intitle:Calibration+of+the+6+DOF+High-Precision+Flexure+Parallel+Robot+?Sigma+6?#0

[20] J. C. Ziegert and P. Kalle, "Error compensation in machine tools: a neural network approach," *J. Intell. Manuf.*, vol. 5, no. 3, pp. 143–151, 1994, doi: 10.1007/BF00123919.

[21] V. Schmidt, B. Müller, and A. Pott, "Solving the forward kinematics of cable-driven

parallel robots with neural networks and interval arithmetic," *Mech. Mach. Sci.*, vol. 15, pp. 103–110, 2014, doi: 10.1007/978-94-007-7214-4_12.

[22] A. Morell, L. Acosta, and J. Toledo, "An artificial intelligence approach to forward kinematics of Stewart Platforms," *2012 20th Mediterr. Conf. Control Autom. MED 2012 - Conf. Proc.*, pp. 433–438, 2012, doi: 10.1109/MED.2012.6265676.

[23] Y. K. Yiu, J. Meng, and Z. X. Li, "Auto-calibration for a parallel manipulator with sensor redundancy," *Proc. - IEEE Int. Conf. Robot. Autom.*, vol. 3, pp. 3660–3665, 2003, doi: 10.1109/robot.2003.1242158.

[24] J. Michaloski, "Coordinated Joint Motion Control for an Industrial Robot," no. March, 1988.

[25] H. Zhuang, L. Liu, and O. Masory, "Autonomous calibration of hexapod machine tools," *J. Manuf. Sci. Eng. Trans. ASME*, vol. 122, no. 1, pp. 140–148, 2000, doi: 10.1115/1.538893.

[26] L. Tsai, "Robot analysis: the mechanics of serial and parallel manipulators," *The Mechanics of Serial and Parallel Manipulators*. p. 520, 1999.

[27] T. Charters, R. Enguiça, and P. Freitas, "Detecting Singularities of Stewart Platforms," *Ind. Case Stud. J.*, vol. 1, pp. 66–80, 2009, [Online]. Available: https://cdn.instructables.com/ORIG/FFK/LAIV/I55MRG6M/FFKLAIVI55MRG6M.pdf

[28] B. Li, Y. Cao, Q. Zhang, and Z. Huang, "Position-singularity analysis of a special class of the Stewart parallel mechanisms with two dissimilar semi-symmetrical hexagons," *Robotica*, vol. 31, no. 1, pp. 123–136, 2013, doi: 10.1017/S0263574712000148.

[29] J. Denavit and R. S. Hartenberg, "A Kinematic Notation for Lower-Pair Mechanisms Based on Matrices," *Journal of Applied Mechanics*, vol. 22, no. 2. pp. 215–221, 1955. doi:

10.1115/1.4011045.

[30] M. Granja, N. Chang, V. Granja, M. Duque, and F. Llulluna, "Comparison between standard and modified Denavit-Hartenberg Methods in Robotics Modelling," *Proc. World Congr. Mech. Chem. Mater. Eng.*, vol. 1, no. 1, pp. 1–10, 2016, doi: 10.11159/icmie16.118.

[31] John J. Craig, "Introduction to Robotics: Mechanics and Control," *Pearson Prentice Hall*, no. 3e, p. 408, 2004.